\def\cvprPaperID{7268}
\def\confName{CVPR}
\def\confYear{2023}

\newcommand*{\affmark}[1][*]{\textsuperscript{#1}}

\renewcommand{\thefootnote}{\fnsymbol{footnote}}

\def\authorBlock{
    Youxin Pang\affmark[1,2]\footnotemark[2]  \qquad
    Yong Zhang\textsuperscript{3}\footnotemark[1] \qquad
    Weize Quan\textsuperscript{1,2} \qquad 
    Yanbo Fan\textsuperscript{3} \qquad 
    Xiaodong Cun\textsuperscript{3} \qquad \\
    Ying Shan\textsuperscript{3} \qquad
    Dong-ming Yan\textsuperscript{1,2}\footnotemark[1] \qquad\\
    \textsuperscript{1}NLPR, Institute of Automation, Chinese Academy of Sciences \\
    \textsuperscript{2}School of Artificial Intelligence, University of Chinese Academy of Sciences\\
    \textsuperscript{3}Tencent AI Lab, ShenZhen, P.R.China \\
}

\newif\ifreview \newcommand{\review}{\reviewtrue}
\newif\ifarxiv 
\newif\ifcamera 
\newif\ifrebuttal

\review 

\pdfoutput=1
\documentclass[10pt,twocolumn,letterpaper]{article}
\usepackage{cvpr}
\usepackage{graphicx}
\usepackage{amsmath}
\usepackage{amssymb}
\usepackage{booktabs}
\usepackage{float}

\usepackage{multirow} 
\usepackage{multicol} 
\usepackage{fancyhdr}

\usepackage{times}
\usepackage{microtype}
\usepackage{epsfig}
\usepackage[table,xcdraw]{xcolor}
\usepackage{caption}
\usepackage{float}
\usepackage{placeins}
\usepackage{color, colortbl}
\usepackage{stfloats}
\usepackage{enumitem}
\usepackage{tabularx}
\usepackage{xstring}
\usepackage{multirow}
\usepackage{xspace}
\usepackage{url}
\usepackage{subcaption}
\usepackage{xcolor}
\usepackage{bm}
\usepackage[hang,flushmargin]{footmisc}
\usepackage{graphicx}

\ifcamera \usepackage[accsupp]{axessibility} \fi

% ...

\ifarxiv  \fi

\newcommand{\R}[1]{{%
    \textbf{%
        \ifstrequal{#1}{1}{\textcolor{red}{R#1}}{%
        \ifstrequal{#1}{2}{\textcolor{blue}{R#1}}{%
        \ifstrequal{#1}{3}{\textcolor{magenta}{R#1}}{%
        \ifstrequal{#1}{4}{\textcolor{teal}{R#1}}{%
                           \textcolor{cyan}{R#1}%
        }}}}%
    }%
}}

\usepackage{xr-hyper}

\makeatletter
\newcommand*{\addFileDependency}[1]{
  \typeout{(#1)}
  \@addtofilelist{#1}
  \IfFileExists{#1}{}{\typeout{No file #1.}}
}

\makeatother

\usepackage[pagebackref,breaklinks,colorlinks]{hyperref}
\usepackage[capitalize]{cleveref}
\crefname{section}{Sec.}{Secs.}
\crefname{table}{Table}{Tables}
\crefname{figure}{Fig.}{Figs.}

\begin{document}
\title{DPE: Disentanglement of Pose and Expression for General Video Portrait Editing}
\author{\authorBlock}

\twocolumn[{
\maketitle

\begin{center}
    \captionsetup{type=figure}
    \includegraphics[width=0.95\textwidth]{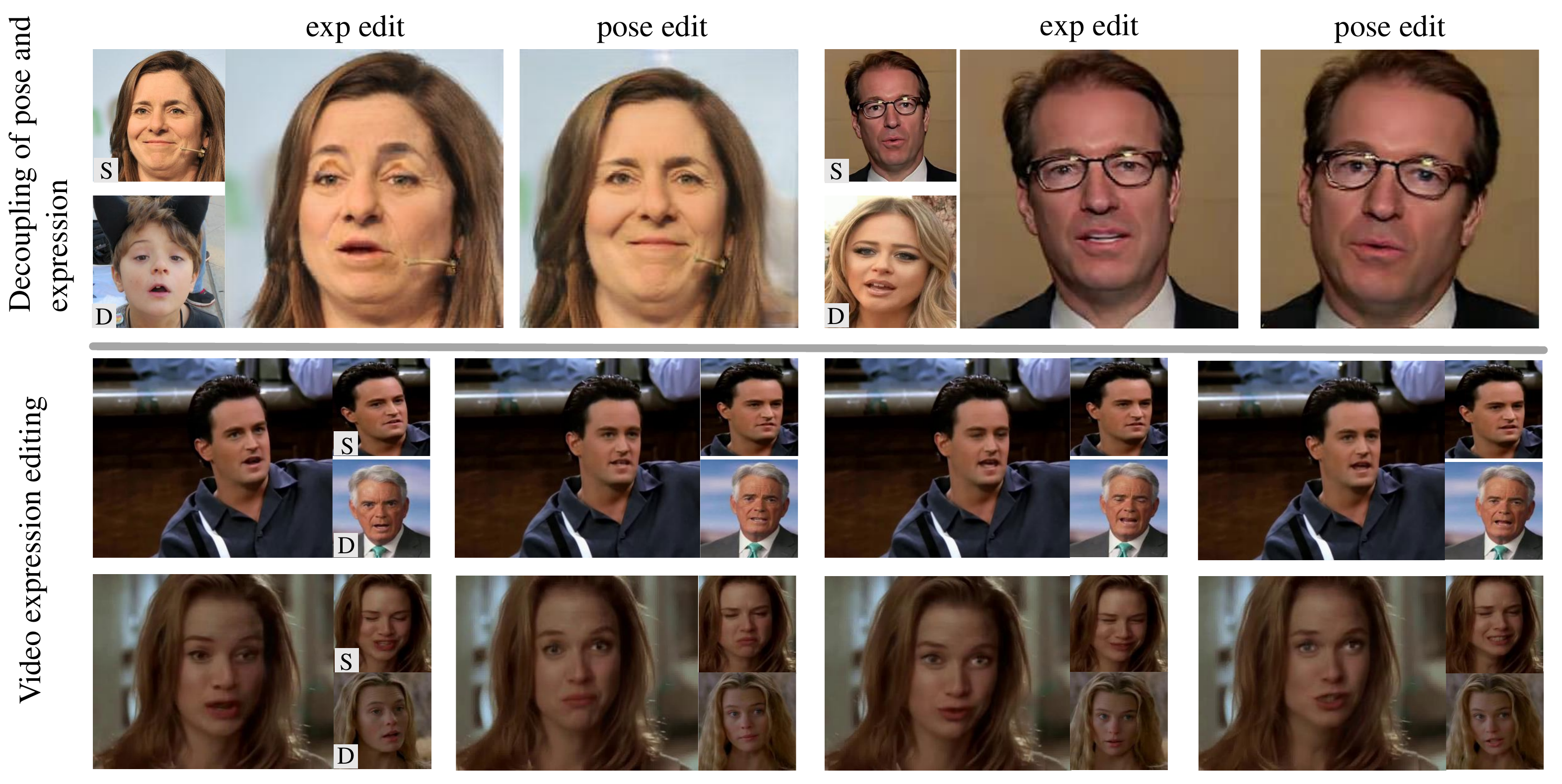}
    \captionof{figure}{ Visual examples produced by our method. 
    Top: disentanglement of pose and expression. Bottom: general video editing. 
    Our method can edit the pose or expression of the source image independently according to the driving image through decoupling pose and expression in motion transfer. 
    Benefiting from the disentanglement, our one-shot talking face method can be applied to video portrait editing. 
    Since our method can edit expression only, the edited cropped face can be pasted back to the full image simply. 
    As our method is subject-agnostic, it can be used to edit any unseen video as well, which is different from subject-dependent video editing methods such as DVP~\cite{kim2018deep}. 
    }
\end{center}
}]

\renewcommand{\thefootnote}{\fnsymbol{footnote}}
\footnotetext[2]{Work done during an internship at Tencent AI Lab.}
\footnotetext[1]{Corresponding Authors.}

\begin{abstract}

One-shot video-driven talking face generation aims at producing a synthetic talking video by  transferring the facial motion from a video to an arbitrary portrait image. 
Head pose and facial expression are always entangled in facial motion and transferred simultaneously.
However, the entanglement sets up a barrier for these methods to be used in video portrait editing directly, where it may require to modify the expression only while maintaining the pose unchanged. 
One challenge of decoupling pose and expression is the lack of paired data, such as the same pose but different expressions.
Only a few methods attempt to tackle this challenge with the feat of 3D Morphable Models (3DMMs) for explicit disentanglement. 
But 3DMMs are not accurate enough to capture facial details due to the limited number of Blenshapes, which has side effects on motion transfer. 
In this paper, we introduce a novel self-supervised disentanglement framework to decouple pose and expression without 3DMMs and paired data, which consists of a motion editing module, a pose generator, and an expression generator. 
The editing module projects faces into a latent space where pose motion and expression motion can be disentangled, and the pose or expression transfer can be performed in the latent space conveniently via addition. 
The two generators render the modified latent codes to images, respectively. 
Moreover, to guarantee the disentanglement, we propose a bidirectional cyclic training strategy with well-designed constraints.  
Evaluations demonstrate our method can control pose or expression independently and be used for general video editing.

\end{abstract}

\section{Introduction}
\label{sec:intro}
Talking face generation has seen tremendous progress in visual quality and accuracy over recent years. 
Literature can be categorized into two groups, \ie, audio-driven~\cite{10.1145/3394171.3413532} and video-driven~\cite{hong2022depth}. 
The former focuses on animating an unseen portrait image or video with a given audio. 
The latter aims at animating with a given video. 
Talking face generation has a variety of meaningful applications, such as digital human animation, film dubbing, etc.
In this work, we target video-driven talking face generation. 


Recently, most methods~\cite{hong2022depth,yao2021one,wang2022latent,Siarohin_2019_NeurIPS} endeavor to drive a still portrait image with a video from different perspectives, \ie, one-shot talking face generation.
But only a few~\cite{kim2018deep, thies2016face2face,ma2019real} make effort to reenact the portrait in a video with another talking video, \ie, video portrait editing. 
This is a more challenging task because edited faces are required to paste back to the original video and temporal dynamics need to be maintained. 
Several methods~\cite{kim2018deep, sun2022continuously} provide personalized solutions to this challenge by training a model on the videos of a specific person only. 
However, the learned model cannot generalize to other identities as the personalized training heavily overfits the facial motion of the specific person and the background.   
For general video portrait editing, therefore, resorting to the generalization property of one-shot talking face generation might be a feasible solution. 

One-shot methods can transfer facial motion from a driving face to a source one, resulting in that the edited face mimics the head pose and facial expression\footnote{Note that facial expression here differs from emotion.} of the driving one.  
The facial motion consists of entangled pose motion and expression motion, which are always transferred simultaneously in previous methods. 
However, the entanglement makes those methods unable to transfer pose or expression independently. 
Since the input to the processing network is always the cropped face rather than the full original image, if the pose is modified along with the expression, the paste-back operation can cause noticeable artifacts around the crop boundary, \eg, twisted neck and inconsistent background.
Consequently, most one-shot methods face this obstacle preventing their application to general video portrait editing.

One challenge to disentangle pose and expression is the lack of paired data, such as the same pose but different expressions, or vice versa. 
In the literature, there are only a few exceptions that can get rid of this limitation, \eg, PIRenderer~\cite{ren2021pirenderer} and StyleHEAT~\cite{2203.04036}, which are based on 3D Morphable Models (3DMMs)~\cite{blanz1999a}, a predefined parametric representation that decomposes expression, pose, and identity. 
However, the 3DMM-based methods heavily depend on the decoupling accuracy of the 3DMM parameters, which is far from satisfactory to reconstruct facial details due to the limited number of Blendshapes. 
Besides, optimization-based 3DMM parameter estimation is not efficient while learning-based estimation will introduce more errors.

In this work, we propose a novel self-supervised disentanglement framework to decouple pose and expression, breaking through the limitation of paired data without using 3DMMs.  
Our framework has a motion editing module, a pose generator, and an expression generator. 
The editing module projects faces into a latent space where coupled pose and expression motion in a latent code can be disentangled by a network. 
Then, pose or expression transfer can be performed by directly adding the latent code of a source face with the disentangled pose or expression motion code of a driving face. 
Finally, the two generators render modified latent codes to images.  
More importantly, to accomplish the disentanglement without paired data, we introduce a bidirectional cyclic training method with well-designed constraints. 
Specifically, given a source face $S$ and a driving one $D$, we transfer the \textit{expression and pose} from $D$ to $S$ sequentially, resulting in two synthetic faces, $S'$ and $S''$. 
Since there is no paired data, no supervision is provided for $S'$. 
To tackle the missing supervision, we exchange the role of $S$ and $D$ to transfer the \textit{pose and expression} motion from $S$ to $D$, resulting in $D'$ and $D''$. 
The distance between $D'$ and $S'$ is one constraint for learning. 
However, it is still not enough for disentangling pose and expression. 
Then, we discover another core constraint, \textit{i.e.,} face reconstruction. 
When $S$ and $D$ are the same, $S'$ and $D'$ are exactly the same as $S$ and $D$, respectively. 
More analyses will be presented in Sec.~\ref{sec:method}.


Our main contributions are three-fold:
\begin{itemize}
    \item We propose a self-supervised disentanglement framework to decouple pose and expression for independent motion transfer, without using 3DMMs and paired data.   
    
    \item We propose a bidirectional cyclic training strategy with well-designed constraints to achieve the disentanglement of pose and expression. 
    
    
    \item Extensive experiments demonstrate that our method can control pose or expression independently, and can be used for general video editing. 
\end{itemize}
\section{Related Work}
\label{sec:related}

\subsection{Talking-face Generation}
\textbf{2D-based methods.}
The early works\cite{bansal2018recycle, wu2018reenactgan, wang2018video} are dominated by subject-dependent methods that can only work on a specific person because their models are trained on the video of the specific person. 
Then, several methods\cite{wang2019few, zakharov2019few} attempt to fine-tune a pre-trained model on the data of a target person for individual use. 
Recently, more works focus on learning a one-shot subject-agnostic model\cite{song2018talking, burkov2020neural, chen2019hierarchical, Siarohin_2019_NeurIPS, pumarola2018ganimation, gu2020flnet, ha2020marionette, zakharov2020fast, zakharov2019few,zhao2022thin,cheng2022videoretalking}, \textit{i.e.,} the trained model can be generally applied to an unseen person.
FOMM\cite{Siarohin_2019_NeurIPS} is a representative method that combines motion field estimation and first-order local affine transformations with the help of sparse keypoints. 
After that, Face-vid2vid\cite{wang2021one} makes an improvement to FOMM and learns unsupervised 3D keypoints. 
Similarly, DaGAN\cite{hong2022depth} incorporates depth on the basis of sparse keypoints to ensure geometric consistency. 
LIA\cite{wang2022latent} has the similar formulation of relative motion as FOMM but learns the semantically meaningful directions in latent space instead of using keypoints.
However, these methods can only edit a still portrait since pose and expression are coupled in the facial motion. As the pose is modified along with the expression, the edited cropped face cannot be pasted back to the original image.

\textbf{3D model-based methods.}
Early works\cite{thies2015real, thies2016face2face} usually build a 3D model for a specific person.
Then, a range of approaches focus on using 3D morphable models~\cite{blanz1999a} that explicitly decompose expression, pose, and identity. 
DVP\cite{kim2018deep} extracts 3DMM parameters of the source and target faces, and the face manipulation is achieved by exchanging their 3DMM parameters. 
Based on DVP, NS-PVD\cite{kim2019neural} preserves the target talking style using a recurrent GAN. 
However, these learned models are subject-dependent and cannot generalize.
Recently, more methods\cite{doukas2021headgan, ren2021pirenderer, fried2019text, geng2018warp} target subject-agnostic talking face generation. 
For instance, HeadGAN\cite{doukas2021headgan} uses the rendered image from 3D mesh as the input of the network. 
It presents independent pose and expression editing with manually adjusted parameters but does not show the transfer from another face.  

\subsection{Decoupling}
Several works\cite{wang2021one, Siarohin_2019_NeurIPS, yao2020mesh, wang2022latent, Drobyshev22MP} focus on the detachment of identity-specific and motion-related information to achieve cross-ID driving, but they do not distinguish pose motion from expression motion. 
Only a few works target the disentanglement of pose and expression for talking face generation.
Almost all of them~\cite{doukas2021headgan, ren2021pirenderer,2203.04036} are based on 3DMMs that explicitly decouple pose and expression. 
PIRenderer\cite{ren2021pirenderer} extracts the 3DMM parameters for a driving face through a pre-trained model and then predict the flow given a source face and the 3DMM parameters. 
During inference, it can transfer only the expression from the driving face by replacing the expression parameters of the source face with those of the driving one. 
 StyleHEAT\cite{2203.04036} follows the similar way based on a pre-trained StyleGAN.
However, the performance of these methods heavily depend on the accuracy of 3DMMs. 
3DMMs are known to be not particularly accurate for face reconstruction due to the limited number of Blendshapes.
They have difficulty delineating facial details of face shape, eye, and mouth, which may eventually have side effects on the synthetic results. 
In this work, instead of using 3DMMs, we decompose pose and expression by the proposed self-supervised disentanglement framework with a bidirectional cyclic training strategy.

\section{The Proposed Method}
\label{sec:method}

\begin{figure*}[t]
    \centering
    \includegraphics[width=0.9\linewidth]{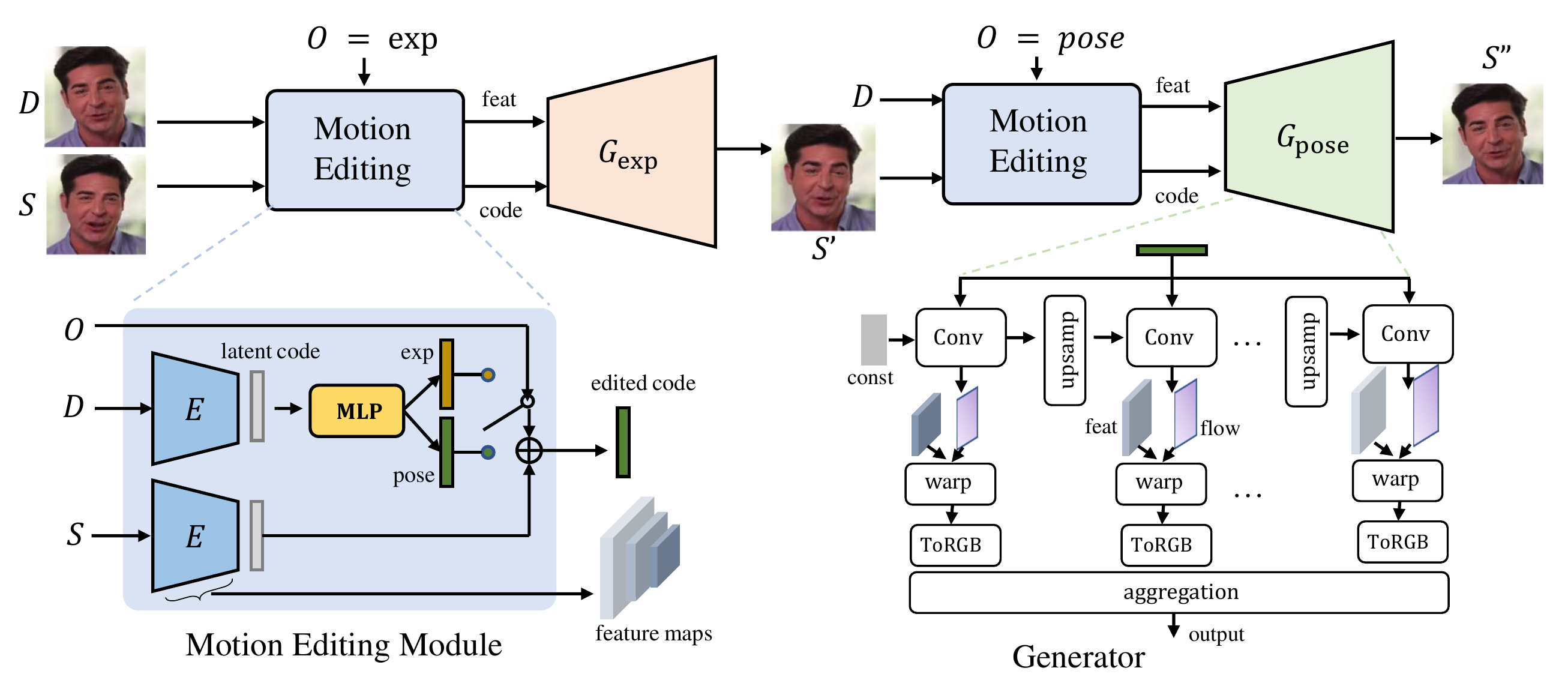}
    \caption{Illustration of our proposed model. The framework consists of three learnable components, \textit{i.e.,} the motion editing module, the expression
generator, and the pose generator. 
    The editing module projects the source and driving images into a latent space where pose motion and expression motion can be disentangled, and then modifies the latent code of the source image according to a given indicator that points out either expression or pose to edit.  
    It outputs an edited latent code and the feature maps of the source image. 
    The pose and expression generators are applied to render the outputs of the editing module to a face image. 
    These two generators share the same architecture but different parameters for interpreting the pose and expression code respectively. }
    \label{fig:network}
\end{figure*}


To apply one-shot talking face generation for general video editing, the disentanglement of pose and expression is indispensable to handle the paste-back operation, \textit{i.e.,} pasting the edited cropped face to the full image. 
In this work, we propose a self-supervised disentanglement framework without paired data and the predefined 3DMMs. 
The whole pipeline is illustrated in Fig.~\ref{fig:network}.  
Our model contains three learnable components, \textit{i.e.,} \textit{the motion editing module}, \textit{the  expression generator}, and \textit{the pose generator}. 
To accomplish the disentanglement, we propose a bidirectional cyclic training strategy to compensate for the missing paired data in which pose or expression are edited individually.
We first introduce the three components in Sec.~\ref{subsec: arc}. 
We then present the training strategy in Sec.~\ref{subsec: training}, followed by the learning objective functions in Sec.~\ref{subsec: losses}.

\subsection{Architecture}~\label{subsec: arc}

\noindent \textbf{Motion Editing Module.}
As shown in Fig.~\ref{fig:network}, given a source image, a driving one, and an editing indicator, the motion editing module yields out an edited latent code and the multi-scale feature maps of the source image. 
The indicator tells either pose or expression of the source image to be edited. 
Inside the module, an encoder is used to project an input image to a latent space that is supposed to be decomposable into two orthogonal subspaces.  
Let $S$, $D$, and $O$ denote the source image, the driving one, and the indicator, respectively. Let $E$ denote the encoder. Then, we have: 
\begin{align}
    \mathbf{c} &= E(X), 
\end{align}
where $X$ is the input of the encoder and $\mathbf{c}$ represents the output latent code. 
$\mathbf{c}_s = E(S)$  and $\mathbf{c}_d = E(D)$ are the latent codes of $S$ and $D$. 

As the driving image provides the facial motion reference, a motion encoder is required to project an image to the same latent space of the encoder. 
Instead of using an separate encoder, we construct the motion space based on the latent space of the encoder. 
Specifically, we use several multiple perceptron (MLP) layers to disentangle the latent space of the encoder to two orthogonal subspaces, \textit{i.e.,} the pose motion space and the expression motion space. 
The architecture of the disentanglement module is that the first few MLP layers act as the shared backbone, followed by two heads that are also composed of MLP layers. 
The disentanglement process can be formulated as: 
\begin{align}
    \mathbf{e}, \mathbf{p} &= \text{MLP}(\mathbf{c}), 
\end{align}
where $\mathbf{e}$ and $\mathbf{p}$ represent the expression and pose motion code, respectively. 
They share the same dimension as $\mathbf{c}$. 

For motion editing, we apply an indicator to specify either pose or expression to edit, which is a binary variable. 
When $O=\text{pose}$, only the pose motion is transferred to the source image. 
When $O= \text{exp}$, the expression is transferred. 
One benefit of disentangling motion in the latent space of the encoder is that motion transfer can be performed by a simple addition, \textit{e.g.,}, the expression editing can be defined as:
\begin{align}
    \overline{\mathbf{c}}_e &= \mathbf{c} + \mathbf{e}, 
\end{align}
where $\overline{\mathbf{c}}_e$ represents the edited code with expression transfer. 
Similar, we have the pose editing, \textit{i.e.,} $\overline{\mathbf{c}}_p = \mathbf{c} + \mathbf{p}$. 

Let $M$ denote the motion editing module. 
The whole process can be defined as: 
\begin{align}
    \overline{\mathbf{c}}, \mathcal{F}  &= M(S, D, O), 
\end{align} 
where $\mathcal{F} = \{\mathbf{F}_k\}^K$ represents the feature maps of the source image, extracted from the encoder. 
$K$ is the number of blocks in the encoder. 
Both the latent code and the feature maps are from the encoder. 
The former represents high-level information while the latter represents mid-level information.

\vspace{1mm}
\noindent \textbf{Pose and Expression Generators.}
The pose or expression of the source image is edited in the latent space by adding the pose or expression motion {from the driving one}. 
Since pose motion captures the global head movement while expression motion captures the local movements of facial components, we use two individual generators for better interpretation of the edited latent code, \textit{i.e.,} the expression generator $G_{e}$ and the pose generator $G_{p}$ . 
The two generators share the same architecture but different parameters. 

Inspired by the flow-based methods~\cite{Siarohin_2019_NeurIPS, ren2021pirenderer}, we use flow fields to manipulate the feature maps. 
Fig.~\ref{fig:network} gives an illustration of the generators. 
Similar to the pipeline of StyleGAN2\cite{Karras2019stylegan2}, we exploit the latent code to generate multi-scale flow fields that are used to warp the feature maps from the encoder in the motion editing module. 
The warped feature maps are aggregated to render an image. 
The expression generator can be defined as: 
\begin{align}
    Y_e &= G_e(\mathbf{c}, \mathcal{F}), 
\end{align} 
where $Y_e$ is the output image of the expression generator.  
Similar, the pose generator is $ Y_p = G_p(\mathbf{c}, \mathcal{F}) $.

\subsection{Bidirectional Cyclic Training Strategy} ~\label{subsec: training}

\begin{figure}[t]
    \centering
    \includegraphics[width=\linewidth]{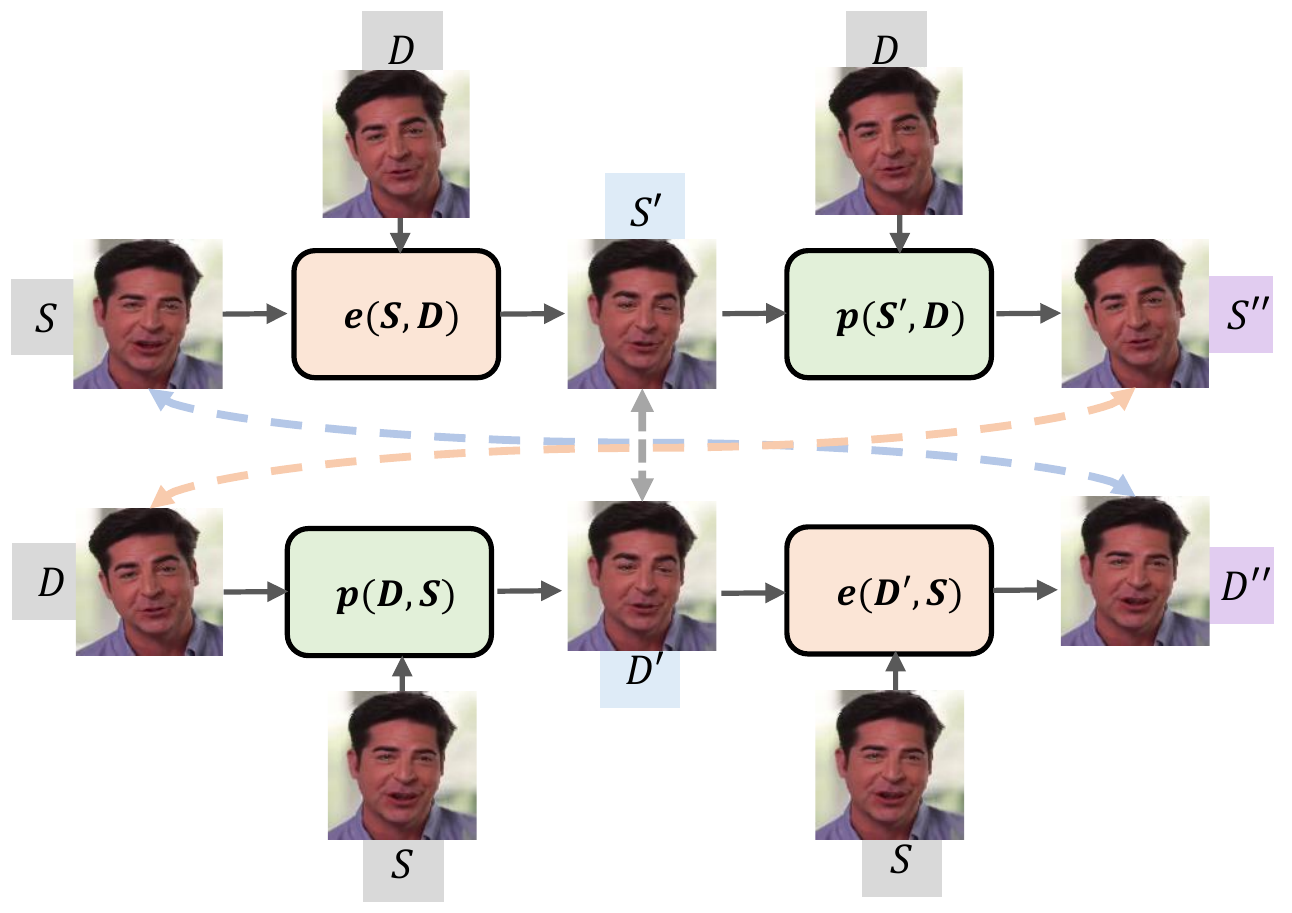}
    \caption{An illustration of the training strategy.}
    \label{fig:training}
\end{figure}
As shown in Fig.~\ref{fig:training},  the pipeline is designed for editing expression and pose independently and sequentially. 
By extracting two frames from a video as input, we can provide supervision at the end of the pipeline. 
However, only such supervision is not enough to disentangle pose and expression. 
Without supervision for the intermediate result (\textit{i.e.,} the output of the expression generator), all the subnetworks will be treated as one network as a whole to complete the reconstruction task with no effort to distinguish the responsibilities of the two generators.

\begin{figure}[t]
    \centering
    \includegraphics[width=\linewidth]{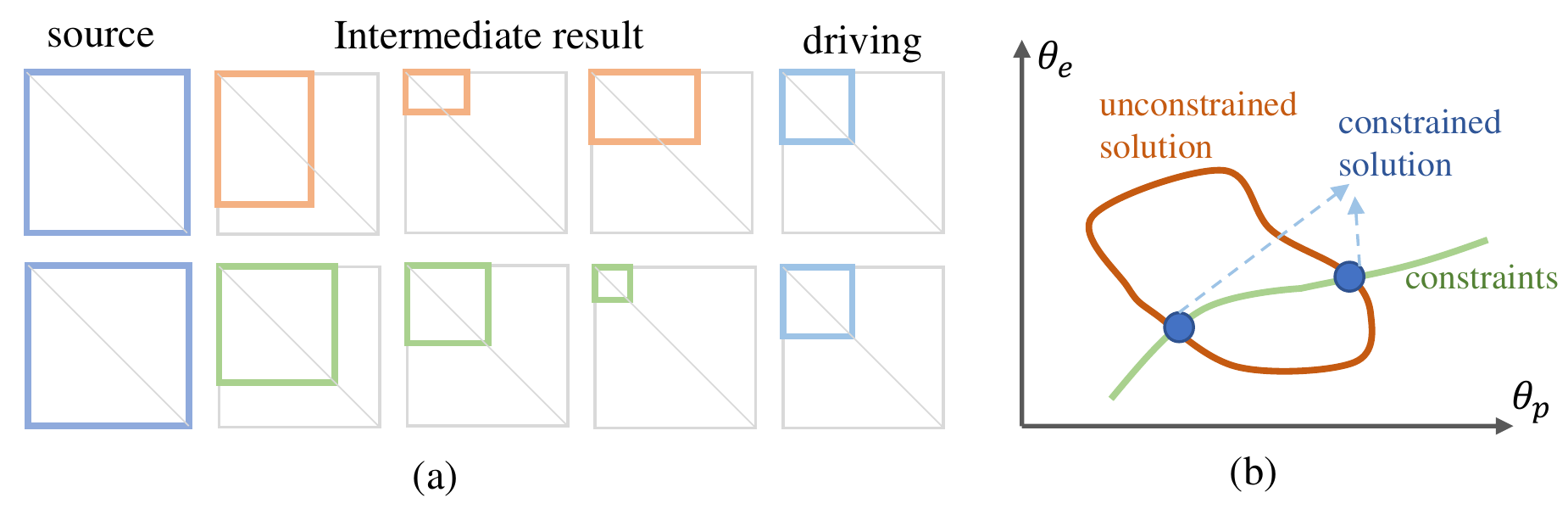}
    \caption{An illustration of the parameter space.}
    \label{fig:params}
\end{figure}
We give a simple illustration in Fig.~\ref{fig:params}(a). 
The task is to scale a large square to a small one in two steps within the range of the gray square. 
Without any constraint, the intermediate result of the first step can be any rectangle in the range (see the top of Fig.~\ref{fig:params}(a)). 
Given the constraint that the height should be the same as the width, the solution space can be greatly narrowed and the intermediate result becomes to be with the property (see the bottom of Fig.~\ref{fig:params}(a)).
Therefore, in our case, given no paired data, we should design a certain constraint to guarantee the disentanglement property of the framework.  
Otherwise, the intermediate face of the expression generator can be in any shape as long as the pose generator can interpret it. 
We further give an illustration from the perspective of the parameter space in Fig.~\ref{fig:params}(b). 
Let $\theta_m$, $\theta_e$, and $\theta_p$ denote the parameters of the motion editing module, the expression generator, and the pose generator, respectively.
For the simplicity of explanation, we assume the motion editing module is optimal, \textit{i.e.,} $\theta_m^*$. 
Without paired data, the solution can be any combination of $\theta_e$ and  $\theta_p$ if they are able to reconstruct the driving image during training. 
If effective constraints are discovered, the solution space can be narrowed and the meaningful solution can be obtained to own the property emphasized by the constraints.

To ensure the disentanglement,
we propose a bidirectional cyclic training strategy without paired data, which is illustrated in Fig.~\ref{fig:training}. 
Let $e(S, D)$ denote expression transfer from $D$ to $S$, \textit{i.e.,} 
\begin{align}
    S' = e(S,D) = G_e(M(S,D,O=\text{exp})), 
\end{align}
where $S'$ is the expression transfer result. 
Let $p(S', D)$ denote pose transfer from $D$ to $S'$, \textit{i.e.,} 
\begin{align}
    S'' = p(S',D) = G_p(M(S',D,O=\text{pose})), 
\end{align}
where $S''$ is the pose transfer result.  
Similarly, we exchange roles of the source and driving images to transfer the pose and expression of the source image to the driving one sequentially. 
Then, we have $D' = p(D, S) $ with pose transferred from $S$, and $D'' = e(D', S)$ with expression from $S$. 

Given tuples $<S, S', S''>$ and $<D, D', D''>$, we can design a set of constraints for the disentanglement. 
As shown in Fig.~\ref{fig:training}, the three dash lines indicate that three pairs of images can be used to compute reconstruction losses, \textit{i.e.,} $<S'', D>$, $<D'', S>$, and $<S', D'>$.  
Please note that though the pair $<S', D'>$ can constrain the intermediate result and narrows the solution space, but it still cannot ensure the disentanglement of pose and expression and the intermediate result is even not face. 

Fortunately, we discover that the self-reconstruction of the two generators is core for the disentanglement, \textit{i.e.,} the pair $<S, e(S, S)>$ and $<S, p(S,S)>$. 
Such pairs encourage the generators to output meaningful face and encourage the editing module to extract the accurate pose and expression motion. 
Otherwise, the generators' outputs will never be the same as the input and there will be always a distance between the two images of a pair.
Theoretically, there is an extreme case where all constraints can be satisfied but the disentanglement fails, \textit{i.e.,} one generator is an identity mapping network while the other generator takes the responsibility of both expression and pose transfer. 
However, we have never met this case in practise. 
One reason is that the generators are randomly initialized and the probability of their convergence to an identity mapping is very low. 

\subsection{Loss Functions} ~\label{subsec: losses}
\textbf{Reconstruction loss $\mathcal{L}_{C}$.}
The \textit{Mean Absolute Error (MAE)} is used to compute the errors between two images in the three pairs:
\begin{equation}
    \mathcal{L}_{rec} = \mathcal{L}_{C}(S'',D) + \mathcal{L}_{C}(D'',S) + \mathcal{L}_{C}(S', D').
\end{equation}

\textbf{Perceptual loss $\mathcal{L}_{P}$.}
To make the synthetic results look more realistic, we also apply the perceptual loss \cite{johnson2016perceptual} to the three pairs as well as the two self-reconstruction pairs:
\begin{equation}
    \begin{aligned}
	    \mathcal{L}_{per} &= \mathcal{L}_{P}(S'', D) + \mathcal{L}_{P}(D'',S) + \mathcal{L}_{P}(S', D')\\
              &+ \mathcal{L}_{P}(e(S,S),S) + \mathcal{L}_{P}(p(S,S),S). \\
    \end{aligned}
\end{equation}

\textbf{Expression loss $\mathcal{L}_{E}$.}
To help with the disentanglement of pose and expression, inspired by spectre\cite{filntisis2022visual}, an expression recognition network\cite{DECA:Siggraph2021} is utilized to obtain the feature vectors. Then we minimize the distance between the feature vectors of the ground-truth and intermediate synthetic images:
\begin{equation}
    \mathcal{L}_{exp} = \mathcal{L}_{E}(S', D) + \mathcal{L}_{E}(D', D). 
\end{equation}

\textbf{GAN loss $\mathcal{L}_{G}$.}
We adopt the non-saturating adversarial loss as our adversarial loss. We also use a discriminator to distinguish reconstructed images from the original ones:
\begin{equation}
    \mathcal{L}_{adv} = \mathcal{L}_{G}(S'') + \mathcal{L}_{G}(D''). 
\end{equation}

Overall, the full objective function is defined as:
\begin{equation}
    \mathcal{L} = \mathcal{L}_{rec} + \lambda_{p}\mathcal{L}_{per} + \lambda_{e}\mathcal{L}_{exp} + \mathcal{L}_{adv}, 
    \label{eq:first}
\end{equation}
where $\lambda_{p}$ and $\lambda_{e}$ are the trade-off hyper-parameters. 


\section{Experiments}
\label{sec:expe}

\begin{figure*}[t]
    \centering
    \includegraphics[width=\linewidth]{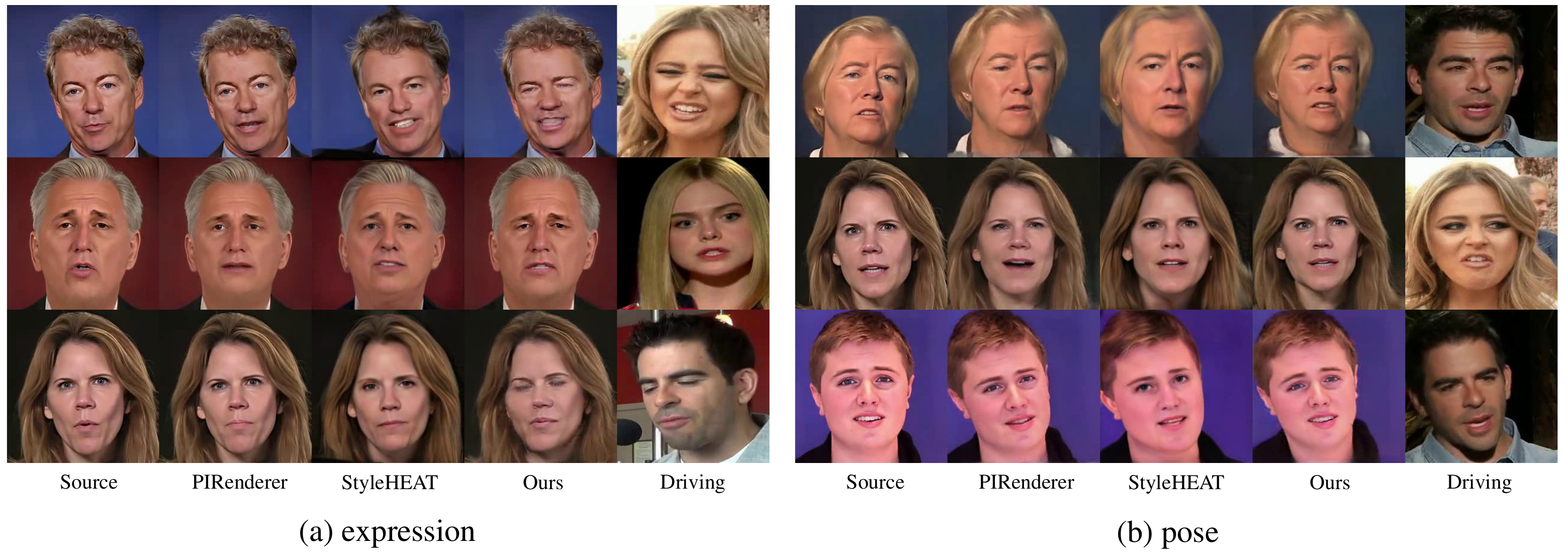}
    \caption{Visual comparisons of independent editing of pose and expression.}
    \label{fig:independent}
\end{figure*}
\subsection{Settings}
\textbf{Datasets.}
We train our model on the VoxCeleb dataset\cite{nagrani2017voxceleb} that includes over 100K videos of 1,251 subjects. 
Following \cite{Siarohin_2019_NeurIPS}, we crop faces from the videos and resize them to $256 \times 256$. 
Faces move freely within a fixed bounding box and no need to align.
For evaluation, the test set contains videos from the VoxCeleb dataset and the HDTF dataset \cite{zhang2021flow}, which are unseen during training.
We collect 15 image-video pairs of different identities from the test set. 
For same-identity reenactment, we use the first frame as the source image and the last 400 frames as the driving images. 
For cross-identity reenactment, we use the first 400 video frames to drive the image in each image-video pair. 
Hence, we can obtain 6K synthetic images for each method for evaluation. 

\textbf{Metrics.}
We utilize a range of metrics to evaluate image
quality and motion transfer quality. 
For same-identity evaluation, peak signal-to-noise ratio (PSNR), structural similarity index (SSIM), and learned perceptual image patch similarity (LPIPS) \cite{zhang2018unreasonable} are used to measure the reconstruction quality. 
And the cosine similarity (CSIM) of identity embedding is used to measure identity preservation.
{For cross-identity and video portrait editing evaluation, AED and APD from PIRender~\cite{ren2021pirenderer} are used to calculate the average 3DMM expression and pose distance between the generated images and targets respectively. }

\textbf{Implementation details.}
We train the model in two stages. 
In the first stage, the three components are jointly optimized for 100K iterations. 
As the expression motion captures local details of facial components, the expression generator is more difficult to learn than the pose generator.
Hence, in the second stage, we learn the expression generator for 50K iterations with fixing the parameters except those of MLPs in the motion editing module and the parameters of the pose generator.  
We set $\lambda_{p}=20$ and $\lambda_{e}=20$. 
The batch size is 32. Adam~\cite{kingma2015adam} is selected as the optimizer with the learning rate of $0.002$ for the first stage and $0.0008$ for the second one. 
During inference, the two generators can be used independently or jointly with the motion editing module. 
Please refer to the supplementary for more details. 

\begin{table}[t]
\centering
\scalebox{0.7}{
\begin{tabular}{l|l l l| l l l}
\toprule
\multirow{2}*{Method} & \multicolumn{3}{c|}{Same-Identity Reenactment} &  \multicolumn{3}{c}{Cross-Identity Reenactment} \\
\cline{2-7}
 & CSIM $\uparrow$ & AED  $\downarrow$ & APD $\downarrow$ & CSIM $\uparrow$ & AED $\downarrow$ & APD $\downarrow$ \\
\toprule


PIRenderer\cite{ren2021pirenderer} & 0.9075 & 0.1205 & \textbf{0.01254} & 0.9133 & 0.2674 & \textbf{0.01182} \\
StyleHEAT\cite{2203.04036}& 0.8320 & 0.1511 & 0.01551 & 0.8489 & 0.2701 & 0.01695 \\
Ours & \textbf{0.9091} & \textbf{0.1133} & 0.01720 & \textbf{0.9204} & \textbf{0.2660} & 0.02464 \\
\bottomrule
\end{tabular}
}
\caption{
 Quantitative comparisons on expression editing.
}
\vspace{-0.05in}
\label{tab:vid2vid}
\end{table}

\begin{table}[t]
\centering
\scalebox{0.7}{
\begin{tabular}{ l | l l l | l l l}
\toprule
\multirow{2}*{Method} & \multicolumn{3}{c|}{Same-Identity Reenactment} &  \multicolumn{3}{c}{Cross-Identity Reenactment} \\
\cline{2-7}
 & CSIM $\uparrow$ & AED  $\downarrow$ & APD $\downarrow$ & CSIM $\uparrow$ & AED $\downarrow$ & APD $\downarrow$ \\
\toprule
PIRenderer\cite{ren2021pirenderer} & 0.9055 & 0.0972 & \textbf{0.01718} & 0.8406 & 0.1397 & \textbf{0.02533} \\
StyleHEAT\cite{2203.04036}& 0.8358 & 0.1285 & 0.02975 & 0.8058 & 0.1577 & 0.03025 \\
Ours & \textbf{0.9192} & \textbf{0.0807} & 0.02459 & \textbf{0.8798} & \textbf{0.1250} & 0.03630\\
\bottomrule
\end{tabular}
}
\caption{
 Quantitative comparisons on pose editing.
}
\vspace{-0.05in}
\label{tab:vid2vid1}
\end{table}
\subsection{Disentanglement for Video Portrait Editing}
Only a few one-shot talking head methods can edit expression or pose independently and be applicable to general video portrait editing.  
Their disentanglement are almost based on the pre-defined 3DMMs while our method is a self-supervised disentanglement without using 3DMMs. 
We compare with two state-of-the-art methods that are open-sourced, \textit{i.e.,} PIRender~\cite{ren2021pirenderer} and StyleHEAT~\cite{2203.04036}. 
Since they use the 3DMM parameters as an input to generate warping flow, they perform independent editing by replacing the pose or expression parameters of the source with those of the driving one. 
Differently, our independent editing is performed in the latent space by simply adding the motion code extracted from the driving image to the latent code of the source one.




\textbf{Qualitative Evaluation.} The visual comparisons are shown in Fig.~\ref{fig:independent}. 
The analyses are summarized as follows. 
First, our method achieves better accuracy in expression transfer than the other two methods, especially the eyes and the mouth shape (see Fig.~\ref{fig:independent}(a)). 
For instance, as shown in the third row, the eyes of the face synthesized by PIRender are `\texttt{open up}' while those of the driving image are `\texttt{closed}'. 
Our method preserves the eye status better. 
In the first and second row, our method captures better mouth movement. 
The reason is that the extracted 3DMM parameters by a pre-trained network cannot accurately reflect the status of eyes and mouth due to the limited number of Blendshapes. 
This is the common weakness of 3DMM-based methods. 
Differently, our method projects images into a latent space with no dependence on 3DMMs, which can depict more accurate facial motions.  

Second, our method preserves the identity better than other two methods in pose transfer (see Fig.~\ref{fig:independent}(b)).
It can be observed that PIRender and StyleHEAT tend to change the face shape of the source image if the face shape of the driving image differs from the source. 
The shape of the synthetic face becomes similar to the driving one. 
For instance, the synthetic image of PIRender in the second row has a wider face than the source while the cheek of the synthetic image in the first row becomes sharper. This might be caused by the incomplete disentanglement of shape, expression, and identity in the 3DMM parameters.  
Unlike them, our method is based on the disentanglement of pose and expression in the latent space, and the pose and expression are decoupled better with the self-supervised learning framework.  
Besides, the teeth generated by our method have better quality than other methods. Though StyleHEAT can generate high-resolution images, the teeth always have artifacts due to the imperfect ruling of StyleGAN. 
While PIRender produces blurry teeth because no parameters in 3DMMs are used to depict the teeth. 
Moreover, it can be observed that StyleHEAT has the identity loss because of the GAN inversion process.

\begin{figure}[t]
    \centering
    \includegraphics[width=\linewidth]{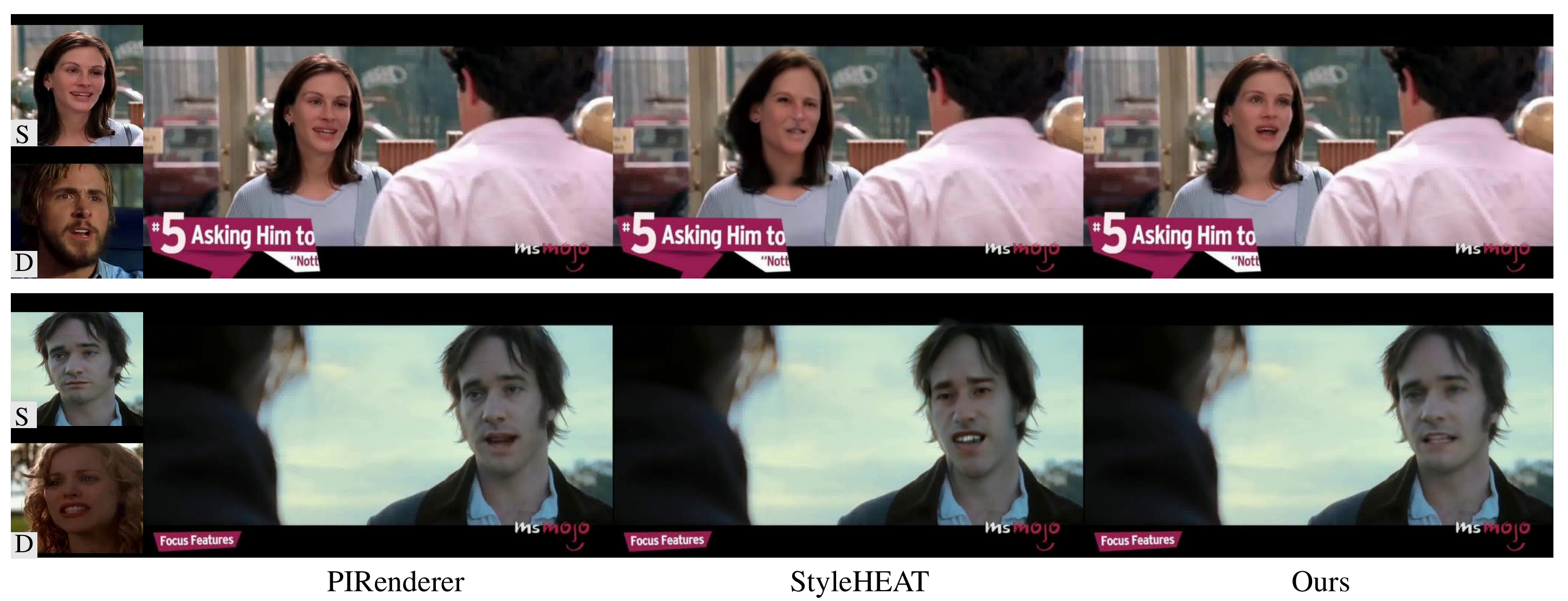}
    \caption{Qualitative comparisons for video expression editing.
    }
    \label{fig:exp_paste}
\end{figure}

\textbf{Quantitative Evaluation.} 
The quantitative comparisons of expression and pose editing are shown in Tab.~\ref{tab:vid2vid} and Tab.~\ref{tab:vid2vid1}, respectively. 
It can be observed that our method achieves better performance in identity and expression preservation in all testing scenarios. 
These results are consistent with the observations in the visual results. 
However, our performance in pose preservation is slightly worse than PIRender. 
Pose transfer reflects the global head movement while expression transfer reflects the local movement of facial components. 
The 3DMM parameters are enough to depict the global head motion but unable to captures the local subtle motions due to the detail level of the Blenshapes. 

\textbf{Video Portrait Editing.} 
The obstacle of applying one-shot talking face generation method to video expression editing is the paste-back operation, \textit{i.e.,} pasting the edited cropped image to the full image. 
If the pose is changed, the edited image cannot be pasted back anymore. 
Benefiting from the disentanglement of pose and expression, only methods that can edit expression independently can be used for video editing. 
Fig.~\ref{fig:exp_paste} illustrates the comparisons between our method and other methods. Our method achieves the better visual quality. 
For these methods, the edited face is blended into the full image with a simple Gaussian blur on the boundaries. 
More videos are provided in the supplementary.

\subsection{One-shot Talking Face Generation}
Our pose and expression generator can be used jointly to transfer both pose and expression from a driving image to a source one. 
Hence, we also compare with several state-of-the-art methods that can only edit pose and expression simultaneously. 
The competing methods are FOMM~\cite{Siarohin_2019_NeurIPS}, PIRenderer~\cite{ren2021pirenderer}, LIA\cite{wang2022latent}, and DaGAN\cite{hong2022depth}. 
We use their released pre-trained models. 


\begin{figure}[t]
    \centering
    \includegraphics[width=\linewidth]{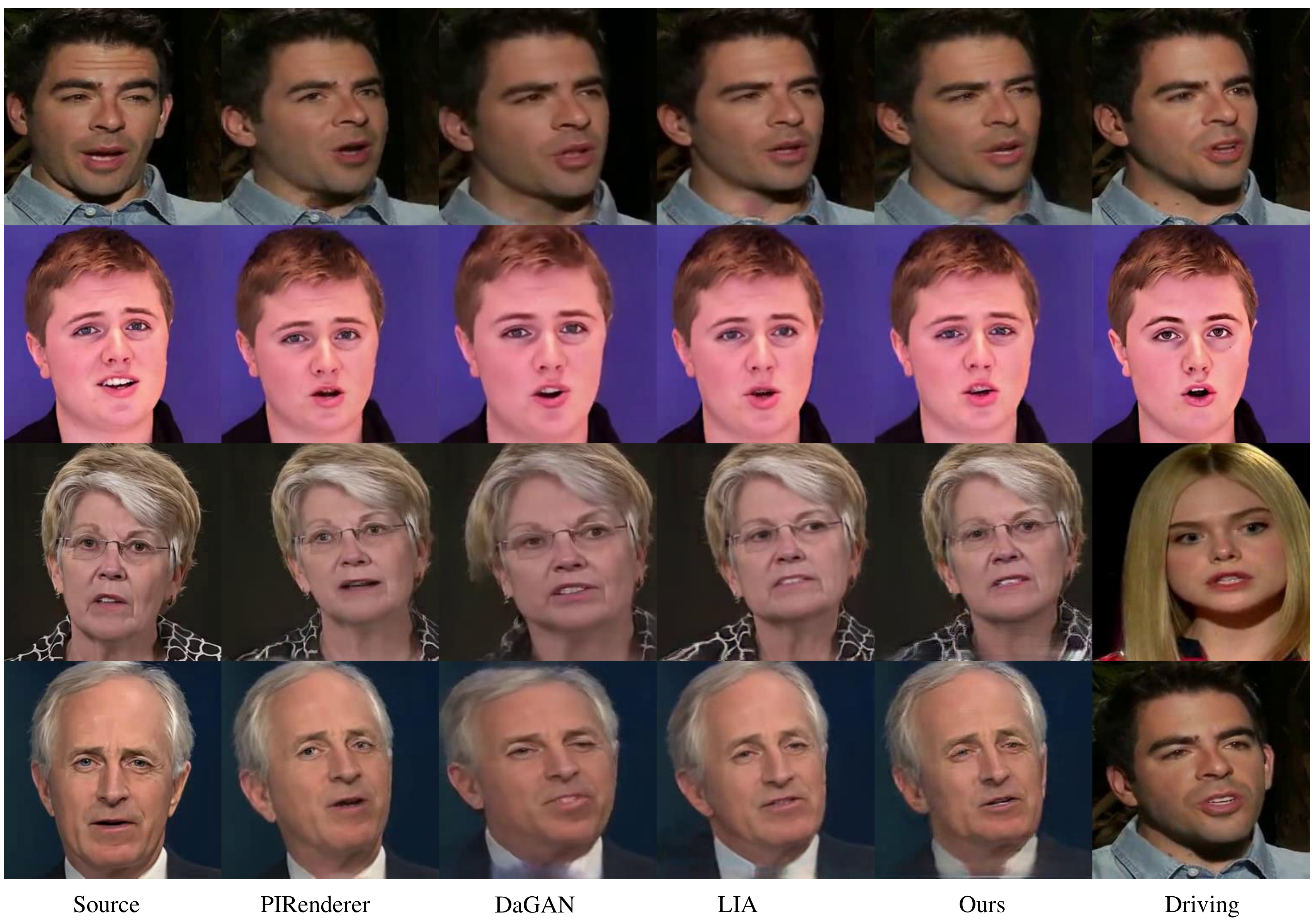}
    \caption{Comparisons with the state-of-the-art methods. }
    \label{fig:sota_oneshot}
\end{figure}

The qualitative comparisons are shown in Fig.~\ref{fig:sota_oneshot}. 
The results of FOMM are reported in the supplementary. 
For same-identity reenactment, our method achieves comparable performance to DaGAN and LIA, and outperforms PIRender. 
PIRender cannot preserve face shape and capture the mouth movement well. 
For cross-identity reenactment, the performance of our method is comparable to LIA and even better in keeping the eye gaze and synthesizing sharp mouth. 
Both LIA and our method are much better than PIRender and DaGAN. 
PIRender produces twisted faces especially when the head pose difference is large between the source and driving faces, while DaGAN changes the face shape a lot and introduces apparent artifacts around the hair. 

The quantitative comparisons are shown in Tab.~\ref{tab:one-shot}. 
For same-identity reenactment, our method achieves the best performance in CSIM, PSNR, SSIM, and APD while our other metrics are comparable to other methods. 
For cross-identity reenactment, our method achieves the best performance in CSIM and APD while AED is comparable to other methods. 

\begin{table*}[t]
\centering

\begin{tabular}{l|l l l l l l|l l l}
\toprule
\multirow{2}*{Method} & \multicolumn{6}{c|}{Same-Identity Reenactment} &  \multicolumn{3}{c}{Cross-Identity Reenactment} \\
\cline{2-10}
 & CSIM $\uparrow$ & LPIPS $\downarrow$ & PSNR $\uparrow$ & SSIM $\uparrow$  & APD  $\downarrow$ & AED $\downarrow$ & CSIM $\uparrow$ & AED $\downarrow$ & APD $\downarrow$ \\
\toprule
FOMM\cite{Siarohin_2019_NeurIPS} & 0.8960 & 0.1536 & 31.1134 & 0.6251 & 0.1000 & 0.01100 & 0.8101 & 0.2570 & 0.02592  \\

PIRender\cite{ren2021pirenderer} & 0.8829 & 0.1713 & 30.7609 & 0.5541 & 0.1110 & 0.01698 & 0.8215 & \textbf{0.2458} & 0.02677
  \\

LIA\cite{wang2022latent} & 0.8906 & \textbf{0.1458} & 31.3371 & 0.6397 & \textbf{0.0998} & 0.01160 & 0.8094 & 0.2659 &  0.02601 \\
DaGAN\cite{hong2022depth} & 0.8910 & 0.1599 & 30.3022 & 0.5904 & 0.1036 &0.01202 & 0.8032 & 0.2584 & 0.02639  \\
Ours & \textbf{0.8965} & 0.1587 & \textbf{31.3631} & \textbf{0.6422} & 0.1000  & \textbf{0.01087} & \textbf{0.8303} & 0.2612 & \textbf{0.02565}\\

\bottomrule
\end{tabular}

\caption{
 Quantitative comparisons with state-of-the-art methods on one-shot talking face generation.
}
\vspace{-0.05in}
\label{tab:one-shot}
\end{table*}

\subsection{Ablation Studies}


\begin{figure}[t]
    \centering
    \includegraphics[width=\linewidth]{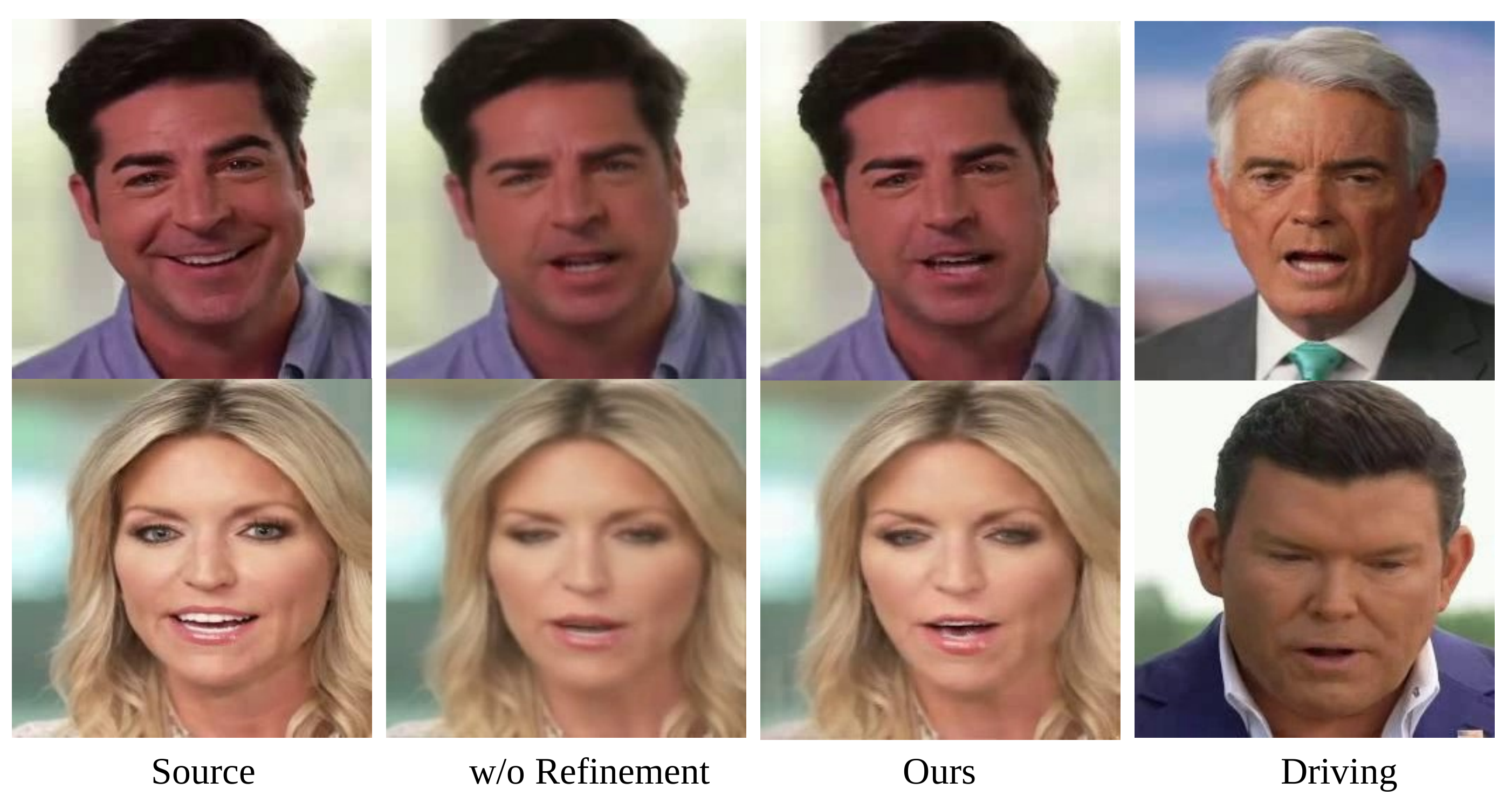}
    \caption{Qualitative ablation studies. The refinement stage helps produce the more realistic images.}
    \label{fig:ablation1}
\end{figure}

\begin{figure}[t]
    \centering
    \includegraphics[width=\linewidth]{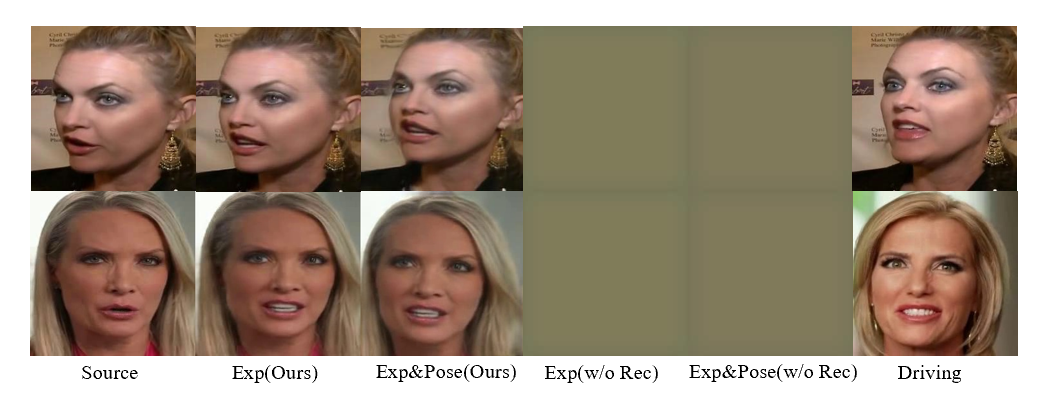}
    \caption{Qualitative ablation studies. The self-reconstruction constraint helps produce the reasonable faces.}
    \label{fig:ablation2}
\end{figure}
\textbf{Refinement Stage.} 
Since pose motion captures the global head movement and expression motion captures the local subtle movement of facial components, we find that expression motion is more difficult to learn than pose motion. 
We fine-tune the expression generator after the joint training of all modules. 
We present the visual improvement of the refinement in Fig.~\ref{fig:ablation1}. 

\textbf{Self-reconstruction Constraint.} 
We reveal that the self-reconstruction constraint for the generators is the core of the disentanglement in the end of Sec.~\ref{subsec: training}. 
We present the intermediate and final results of the forward pass of the framework with or without using the constraint in Fig.~\ref{fig:ablation2}. 
The whole framework is hard to train without the constraint and cannot generate meaningful faces. 

\section{Conclusion}
\label{sec:conclusion}

We propose a novel self-supervised disentanglement framework to decouple pose and expression without 3DMMs and paired data. 
With the powerful editable latent space where pose motion and expression motion can be disentangled, our method can perform pose or expression transfer in this space conveniently via addition.
It enables independent control over pose and expression, and is better than 3DMM in terms of facial expression details with the help of our model.
Benefiting from the disentanglement, our method can be used for general video portrait editing, \textit{i.e,} one model for any unseen person. 
We execute the experiment of video portrait editing, the results demonstrate the advantages of our framework over recent state-of-the-art methods and the application to general video editing. 
We also perform the experiment of one-shot talking face generation, the results show that our method is comparable to other methods.

{\small
\bibliographystyle{ieee_fullname}
\bibliography{11_references}
}


\end{document}